\documentclass[10pt,twocolumn,letterpaper]{article}

\usepackage[pagenumbers]{cvpr} 
\usepackage[accsupp]{axessibility}  

\usepackage{graphicx}
\usepackage{amsmath}
\usepackage{amssymb}
\usepackage{booktabs}
\usepackage[accsupp]{axessibility}
\usepackage{times}
\usepackage{epsfig}
\usepackage{epigraph}
\usepackage{graphicx}
\usepackage{amsmath}
\usepackage{amssymb}
\usepackage[font={small}]{caption}
\usepackage[table]{xcolor}
\usepackage{tabularx}
\usepackage{wrapfig}
\usepackage{multirow}
\usepackage{float}
\usepackage{bm}
\usepackage{adjustbox}
\usepackage{mathtools}
\usepackage{mathtools, cuted}
\usepackage[misc]{ifsym}

\usepackage{makecell}

\definecolor{citecolor}{RGB}{65,105,225}
\usepackage[pagebackref=true,breaklinks=true,letterpaper=true,colorlinks,
citecolor=citecolor,bookmarks=false]{hyperref}

\usepackage[capitalize]{cleveref}
\crefname{section}{Sec.}{Secs.}
\Crefname{section}{Section}{Sections}
\Crefname{table}{Table}{Tables}
\crefname{table}{Tab.}{Tabs.}

\def\cvprPaperID{4725} 
\def\confName{CVPR}
\def\confYear{2023}

\begin{document}

\title{MonoHuman: Animatable Human Neural Field from Monocular Video}

\author{Zhengming Yu$^{1}$, Wei Cheng$^{1,2}$, Xian Liu$^{3}$, Wayne Wu$^{2}$, Kwan-Yee Lin\textsuperscript{2, 3} \textsuperscript{\Letter} \\
\small$^1$SenseTime Research\quad$^2$Shanghai AI Laboratory\\ $^3$\small The Chinese University of Hong Kong\\
{\tt\small 
yuhuaijin36@gmail.com \quad 
chengwei@sensetime.com \quad 
} \\
{\tt\small 
alvinliu@ie.cuhk.edu.hk \quad
wuwenyan0503@gmail.com \quad
junyilin@cuhk.edu.hk}
}


\twocolumn[{
\renewcommand\twocolumn[1][]{#1}
\maketitle
    \vspace*{-8ex}
    \begin{center}
    \includegraphics[width=\textwidth]{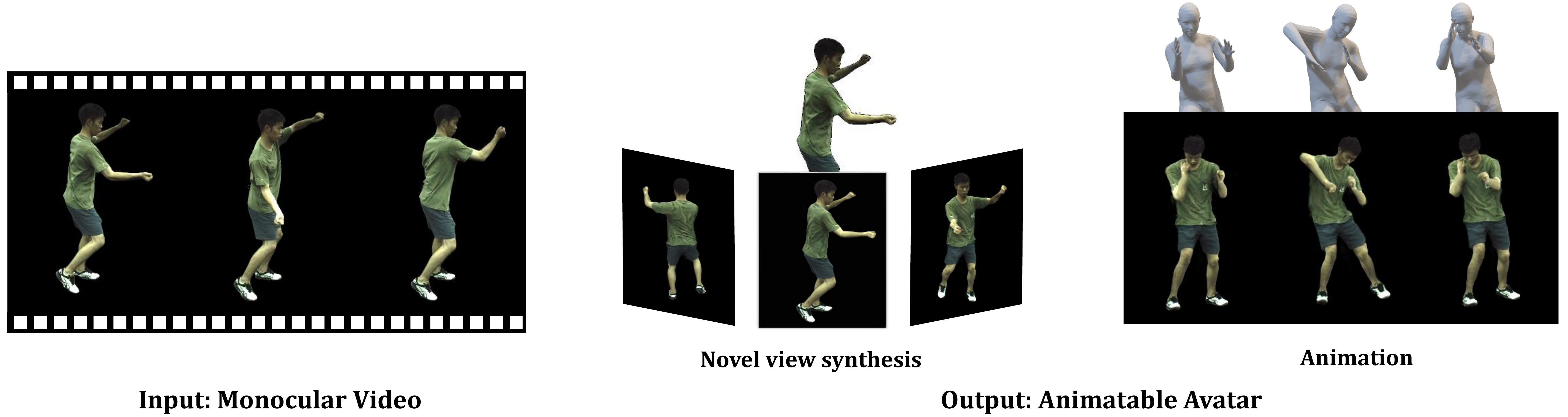}
    \end{center}  
    \vspace*{-20px}
    \captionof{figure}{\textbf{Overview of MonoHuman.} We present MonoHuman, which learns an animatable human neural field from monocular videos. With monocular video input, our method can create an animatable human avatar that supports novel view synthesis and novel pose animation. Our code, models and demos are available at \url{https://yzmblog.github.io/projects/MonoHuman}.}
    \label{fig:teaser}
}
\vspace{2ex}
]

\begin{abstract}
  Animating virtual avatars with free-view control is crucial for various applications like virtual reality and digital entertainment. Previous studies have attempted to utilize the representation power of the neural radiance field (NeRF) to reconstruct the human body from monocular videos. Recent works propose to graft a deformation network into the NeRF to further model the dynamics of the human neural field for animating vivid human motions. However, such pipelines either rely on pose-dependent representations or fall short of motion coherency due to frame-independent optimization, making it difficult to generalize to unseen pose sequences realistically. In this paper, we propose a novel framework \textbf{MonoHuman}, which robustly renders view-consistent and high-fidelity avatars under arbitrary novel poses. Our key insight is to model the deformation field with bi-directional constraints and explicitly leverage the off-the-peg keyframe information to reason the feature correlations for coherent results. Specifically, we first propose a Shared Bidirectional Deformation module, which creates a pose-independent generalizable deformation field by disentangling backward and forward deformation correspondences into shared skeletal motion weight and separate non-rigid motions. Then, we devise a Forward Correspondence Search module, which queries the correspondence feature of keyframes to guide the rendering network. The rendered results are thus multi-view consistent with high fidelity, even under challenging novel pose settings. Extensive experiments demonstrate the superiority of our proposed MonoHuman over state-of-the-art methods.
     
\end{abstract}
\section{Introduction}
\label{sec:intro}

Rendering a free-viewpoint photo-realistic view synthesis of a digital avatar with explicit pose control is an important task that will bring benefits to AR/VR applications, virtual try-on, movie production, telepresence, etc. However, previous methods~\cite{peng2021neural,peng2021animatable,zhao2021humannerf} usually require carefully-collected multi-view videos with complicated systems and controlled studios, which limits the usage in general and personalized scenarios applications. Therefore, though challenging, it has a significant application value to {\textit{directly recover and animate the digital avatar from a monocular video.}}

Previous rendering methods~\cite{peng2021neural} can synthesize realistic novel view images of the human body, but hard to animate the avatar in unseen poses. To address this, some recent methods deform the neural radiance fields (NeRF~\cite{mildenhall2020nerf}) ~\cite{peng2021animatable,weng2022humannerf} to learn a backward skinning weight of parametric models depending on the pose or individual frame index. They can animate the recovered human in novel poses with small variations to the training set. However, as the blending weights are pose-dependent, these methods usually over-fit to the seen poses of training data, therefore, lack of generalizability~\cite{weng2022humannerf}. Notably, one category of approaches~\cite{chen2021snarf, li2022tava} solves this problem by learning a pose-independent forward blend weight in canonical space and using the root-finding algorithm to search the backward correspondence. However, the root-finding algorithm is time-consuming. \cite{chen2022moco} proposes to add a forward mapping network to help the learning of backward mapping, though with consistent constraint, their backward warping weights still are frame dependent. Some other works~\cite{peng2021animatable, jiang2022neuman, te2022neural} leverage the blend weight from the template model like SMPL~\cite{loper2015smpl}. The accuracy of their deformation heavily hinges on the template model and always fails in the cloth parts as SMPL does not model it. How to learn an accurate and generalizable deformation field is still an open problem.

To reconstruct and animate a photo-realistic avatar that can generalize to unseen poses from the monocular video, we conclude three key observations to achieve generalizability from recent studies: $\textbf{1)}$ The deformation weight field should be defined in canonical space and as pose-independent as possible~\cite{chen2021snarf, weng2020vid2actor, weng2022humannerf}; $\textbf{2)}$ An ideal deformation field should unify forward and backward deformation to alleviate ambiguous correspondence on novel poses; $\textbf{3)}$ 
Direct appearance reference from input observation helps improve the fidelity of rendering.

 We propose \textbf{MonoHuman}, a novel framework that enjoys the above strengths to reconstruct an animatable digital avatar from only monocular video. Concretely, we show how to learn more correct and general deformation from such limited video data and how to render realistic results from deformed points in canonical space. 
We first introduce a novel Shared Bidirectional Deformation module to graft into the neural radiance fields, which disentangles the backward and forward deformation into one shared skeletal motion and two separate residual non-rigid motions. The shared motion basis encourages the network to learn more general rigid transformation weights. The separate residual non-rigid motions guarantee the expressiveness of the accurate recovery of pose-dependent deformation. 

With the accurate learned deformation, we further build an observation bank consisting of information from sparse keyframes, and present a Forward Correspondence Search module to search observation correspondence from the bank which helps create high fidelity and natural human appearance, especially the invisible part in the current frame.
With the above designs, our framework can synthesize a human at any viewpoint and any pose with natural shape and appearance.

To summarize, our main contributions are three-fold: 
\begin{itemize}
    \item We present a new approach MonoHuman that can synthesize the free viewpoint and novel pose sequences of a performer with explicit pose control, only requiring a monocular video as supervision. 
    \item We propose the Shared Bidirectional Deformation module to achieve generalizable consistent forward and backward deformation, and the Forward Search Correspondence module to query correspondence appearance features to guide the rendering step. 
    \item Extensive experiments demonstrate that our framework MonoHuman renders high-fidelity results and outperforms state-of-the-art methods.
\end{itemize}

\section{Related Work}\label{sec:related}

\noindent\textbf{Human Performance Capture.} Previous works reconstruct the human body geometry from multi-view videos~\cite{xu2011video,de2008performance,sigal2010humaneva} or depth cameras~\cite{newcombe2015dynamicfusion,shapiro2014rapid,yu2018doublefusion,xu2019flyfusion} and the albedo map of surface mesh~\cite{guo2017real,guo2019relightables}. Recent works model human geometry as radiance fields~\cite{peng2021neural,peng2021animatable,park2021nerfies,park2021hypernerf,kwon2021neural,cheng2022generalizable,zhao2021humannerf,weng2022humannerf,jiang2022neuman} or distance functions~\cite{xu2021h,tiwari2022pose}. NeuralBody~\cite{peng2021animatable} uses structured pose features generated from SMPL~\cite{loper2015smpl} to condition the radiance field, enabling it to recover human performers and produce free-viewpoint images from sparse multi-view videos. Researchers further improve the generalization~\cite{kwon2021neural,cheng2022generalizable,zhao2021humannerf} ability or reconstruction quality~\cite{xu2021h} following the same setup.Although they have improved the rendering quality well, the application scenario of these methods is limited to mostly indoor multi-camera setups with accurate camera calibration and restricted capture volume.

\begin{figure*}[!t]
    \centering
    \vspace{-2ex}
    \includegraphics[width=0.95\linewidth]{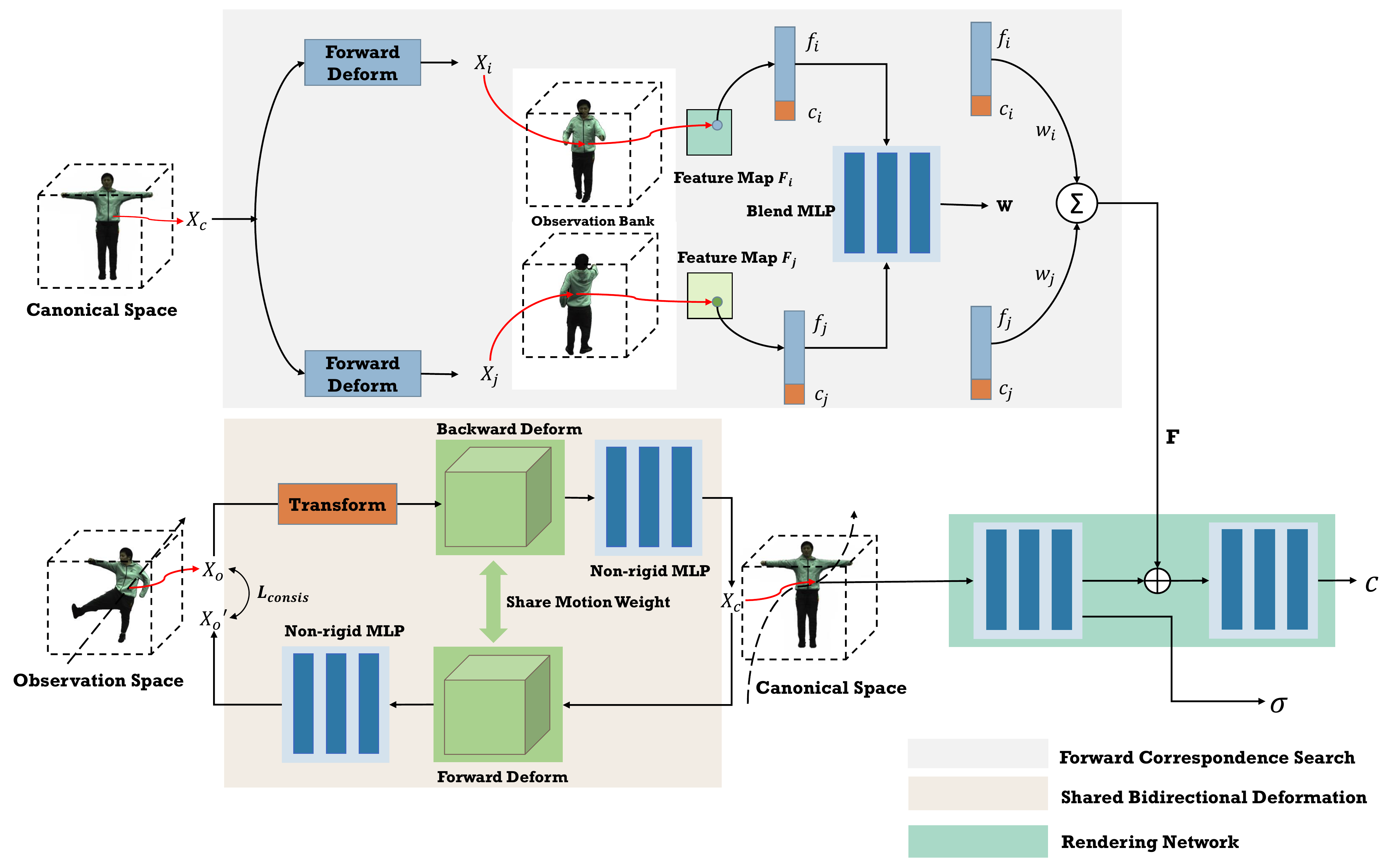}
    \vspace{-2ex}
    \caption{\textbf{Framework of MonoHuman.} In the Shared Bidirectional Deformation module, points in different observation space $\bm{X}_{o}$ will be deformed to $\bm{X}_{c}$ in the same canonical space by backward deforming $\bm{D}_{b}$. And $\bm{X}_{c}$ can be deformed back into the points $\bm{X}^{\mathrm{'}}_o$ in observation space by symmetrical forward deformation $\bm{D}_{f}$. A consistent loss $\bm{L}_{consis}$ is used to regularize the motion weight. In Forward Correspondence Search module, $\bm{X}_{c}$ is deformed to $\bm{X}_{i}$ and $\bm{X}_{j}$ in two correspondence space to get the feature $\bm{f}_{i}$, $\bm{f}_{j}$ and the color $\bm{c}_{i}$ $\bm{c}_{j}$. These features are blended to the feature $\bm{F}$ by the weights Blend MLP generated. $\bm{F}$ is used to guide the rendering network finally.}
    \vspace{-3ex}
    \label{fig:pipeline}
\end{figure*}

\noindent\textbf{Human Rendering from Monocular Video.} To remove the multi-view constraints, many methods explore reconstructing the digital human from a single image or monocular video. \cite{saito2019pifu,saito2020pifuhd,xiu2022icon} recover static clothed 3d human from a single image by learning the accurate surface reconstruction. In order to model the human dynamics, Some works~\cite{xu2018monoperfcap,habermann2020deepcap} learn to deform the pre-scanned human model from monocular video. ~\cite{alldieck2018video} reconstructs the full human body from a self-rotating monocular video by optimizing the displacement of the SMPL model. Recently, SelfRecon~\cite{jiang2022selfrecon} improves the reconstruction quality by representing the body motion as a learnable non-rigid motion along with the pre-defined SMPL skinning. Other works~\cite{park2021nerfies,park2021hypernerf,pumarola2020d} model the dynamic human via deformable NeRF with a deformation or blending network that blends the ray from current motion space to a canonical space. However, the depth ambiguity of monocular video and insufficient pose observation may lead the deformation field to overfit to training data which further causes floating artifacts in novel views. To address these problems, some works~\cite{jiang2022neuman,weng2020vid2actor,weng2022humannerf} introduce motion priors to regularize the deformation. NeuMan~\cite{jiang2022neuman} learns the blending field by extending the deformation weight of the closest correspondence from the SMPL mesh and optimizing it during training. HumanNeRF~\cite{weng2022humannerf} decouples the deformation field as a skeleton motion field and a non-rigid motion field. These methods target at rendering free-viewpoint human images for a bullet-time effect, while ours focuses on animating reconstructed avatars to out-of-distribution poses even challenge poses generated by text prompts.

\noindent\textbf{Human Animation.} Neural Actor~\cite{liu2021neural} proposes deformable NeRF from UV space and refines the detail with the predicted texture map.~\cite{peng2021animatable} create an animatable model by conditioning the NeRF with pose-dependent latent code.~\cite{su2021anerf,2021narf} drive the canonical model by transforming points to local bone coordinates. Moco-Flow~\cite{chen2022moco} uses a complementary time-conditioned forward deformation network with cycle consistency loss to regularize the deformation field. These methods usually lean by memorizing the observed poses and are hard to generalize to novel poses. Recent works~\cite{saito2021scanimate,chen2021snarf} address this problem by combining a forward deformation network with the root-finding algorithm to learn the deformation field. However, the root-finding algorithm they use is highly time-consuming.

\noindent\textbf{Image-based rendering.} 
To synthesize novel view images without recovering detailed 3D geometry, previous works~\cite{gortler1996lumigraph, davis2012unstructured} get novel view images from interpolating the light field. Though achieving realistic results, the view range they can synthesize is limited. To address this problem, some works~\cite{chaurasia2013depth, penner2017soft} seek the help of proxy geometry by inferring the depth map. Although they extend the renderable view range, their generated results are sensitive to the accuracy of the proxy geometry. With the development of deep learning techniques, many works~\cite{choi2019extreme, hedman2018deep, kalantari2016learning, thies2018ignor, wang2021ibrnet, liu2022neural, kwon2020rotationally} introduce learnable components to the image-based rendering methods and improve the robustness. Specifically, IBRNet~\cite{wang2021ibrnet} learns the blending weight of image features from a sparse set of nearby views. NeuRay~\cite{liu2022neural} further predicts the visibility of 3D points to input views within their representation. These works show the powerful effect of image features from different views for rendering.

\section{Our Approach}
We present \textbf{MonoHuman} that reconstruct an animatable digital avatar from monocular video. The generated avatar can be controlled in free view-point and arbitrary novel poses. The pipeline is illustrated in Fig.~\ref{fig:pipeline}. Specifically, we first formulate the problem in Sec.~\ref{sec:3.1}, and then the \emph{Shared Bidirectional Deformation Module} which deforms points between observation space and canonical space is elaborated in Sec.~\ref{sec:3.2}. Sec.~\ref{sec:3.3} introduces the \emph{Forward Correspondence Search Module} to extract the correspondence features in keyframes. Finally, the training objectives and volume rendering process are described in Sec.~\ref{sec:3.4}.

\subsection{Preliminaries and Problem Setting}
\label{sec:3.1}
Given an image sequence of a person and corresponding poses, segmentation masks, as well as the camera's intrinsic and extrinsic parameters, following~\cite{weng2022humannerf, peng2021animatable} we use a neural field to represent the human from observation space in one canonical space. Specifically, the color and density of point ${x_o}$ in observation space are model as:
\begin{equation}
\label{eq:hn_representation}
(\mathbf{c}(\mathbf{x_o}), \mathbf{\sigma}(\mathbf{x_o})) = {F_c}({D}(\mathbf{x_o},\mathbf{p})),
\end{equation}
where ${D}$ is a backward deformation mapping that takes the body pose $\mathit{\mathbf{p}}$ and point in observation space $\mathbf{x_o}$ as input, outputs the point in the canonical space $\mathbf{x_c}$. ${F_c}$ is a mapping network that takes point in the canonical space $\mathbf{x_c}$ as input and outputs its color value c and density $\sigma$. With this representation, the final rendering step using volume rendering~\cite{mildenhall2020nerf,max1995optical} is done in one canonical space.

\subsection{Shared Bidirectional Deformation Module}
\label{sec:3.2}
To avoid the overfitting problem mentioned in Sec.~\ref{sec:intro}, we define the deformation field in canonical space following HumanNeRF~\cite{weng2022humannerf}. However, the proposed single backward deformation is only constrained by the image reconstruction loss, which is under-constrained and will gain more deform errors when it meets unseen poses. We show this in phenomenon Sec.~\ref{sec:experiment}. Intuitively, a loss that is defined only in the deformation field will help to regularize the deformation field. But how to build such a constraint is non-trivial. MoCo-Flow~\cite{chen2022moco} uses an additional time-conditioned forward deformation MLP to constrain the consistency of deformation. BANMo~\cite{Yang_2022_CVPR} uses different MLPs to generate pose-conditioned deformation weights to achieve forward and backward deformation. However, as the deformation fields are two different MLPs and frame-dependent or pose-dependent, they still suffer from the over-fitting problem. Inspired by the works above, we design our Shared Bidirectional Deformation module with a forward and backward deform using the same motion weight defined in canonical space. Since our design is pose-independent, it helps the model’s generalization to out-of-distribution poses. In practice, we formulate the backward deformation as:
\begin{equation} 
    \label{eq:backward-deform-formulate}
D_{b} : (\mathbf{x_o}, \mathbf{p}) \rightarrow \mathbf{x_c}, 
\end{equation}
where $\mathbf{p}$ = $\{{\boldsymbol{\omega}}_i\}$ are K joints rotations represented as axis-angle vectors. Similar to~\cite{weng2022humannerf}, the complete deformation consists of two parts: motion weight deformation and non-rigid deformation:
\begin{equation}
\label{eq:whole-backward-deform}
D_{b}(\mathbf{x_o}, \mathbf{p}) = T^{\text{b}}_{mo}(\mathbf{x_o}, \mathbf{p}) 
  + T^{\text{b}}_{NR}(T^{\text{b}}_{mo}(\mathbf{x_o},\mathbf{p}), \mathbf{p}),
\end{equation}
where computation of ${T^{\text{b}}_{mo}}$ is similar to linear blend skinning:
\begin{equation}
T_{mo}^b(\mathbf{x_o}, \mathbf{p}) = \
  \sum_{i=1}^{K}{w_o^i(\mathbf{x_o}){(R_{i}\mathbf{x_o} + t_{i})}},
\label{eq:motion-deform}  
\end{equation}
in which ${w_o^i}$ is the blend weight of $i$-th bone, and ${R_i}$, $t_i$ are the rotation and translation that map the bone's coordinates from observation to canonical space. We compute ${w_o^i(\mathbf{x_o})}$ by a set of motion weight volumes defined in canonical space ${w_c^i}$: 

\begin{equation}
    {w_o}^i(\mathbf{x_o}) =  \frac{{w_c}^i(R_{i}\mathbf{x_o} + t_{i})}
  {\sum_{k=1}^{K}{{w_c}^k(R_{k}\mathbf{x_o} + t_{k})}}.
\label{eq:observation-weight-computation}  
\end{equation}
We solve the volume ${W_c(x_c)}$ = $\{w_c^i(x_c)\}$ using CNN which generates from a random constant latent code \textbf{z} same as HumanNeRF~\cite{weng2022humannerf}:
\begin{equation}
{W_c({\mathbf{x_c})}} = {\rm CNN}({\mathbf{x_c}}; {\mathbf{z}}).
\label{eq:motion-weight-computation}  
\end{equation}
We additionally use the motion weight defined in canonical space to achieve forward deformation:
\begin{equation} 
    \label{eq:forward-deform-formulate}
D_{f} : (\mathbf{x_c}, \mathbf{p}) \rightarrow \mathbf{x_o}, 
\end{equation}
different from backward deformation, the forward motion weight can be directly queried by ${\mathbf{x_o}}$ as:
\begin{equation}
T_{mo}^f(\mathbf{x_c}, \mathbf{p}) = \
  \sum_{i=1}^{K}{w_c^i(\mathbf{x_c})\mathbf{x_c}},
\label{eq:motion-deform-forward}  
\end{equation}
because the motion weight is defined in canonical space. As for non-rigid deformation, we use another MLP to compute forward deform. Both forward and backward pose-dependent deformation can be written in the same formula:
\begin{equation}
T_{NR}(\mathbf{x}, \mathbf{p}) = \
  {\rm MLP_{NR}}({T_{mo}(\mathbf{x}, \mathbf{p})}, \mathbf{p}),
\label{eq:motion-deform-forward}  
\end{equation}
In order to add the constraint that only related to the deformation field as a regularization, we use the intuition of consistency of forward and backward deformation, the consistent loss $\mathbf{L_{consis}}$ is computed as:
\begin{equation}
{L_{consis}}=
\begin{cases}
 d &\text{if } d \geq \theta \\
 0 &\text{else}
\end{cases} 
 d = L_2({\mathbf{x_o}, {D_{f}(D_{b}(\mathbf{x_o}, \mathbf{p})}}, \mathbf{p})),
\label{eq:loss-consis}  
\end{equation}
where $L_2$ means the $L_2$ distance calculation, and it only penalize the points whose $L_2$ distance is greater than threshold $\theta$ we set to avoid over regularization.

\begin{figure*}[t!]
    \centering
    \vspace{-3mm}
    \includegraphics[width=0.98\linewidth]{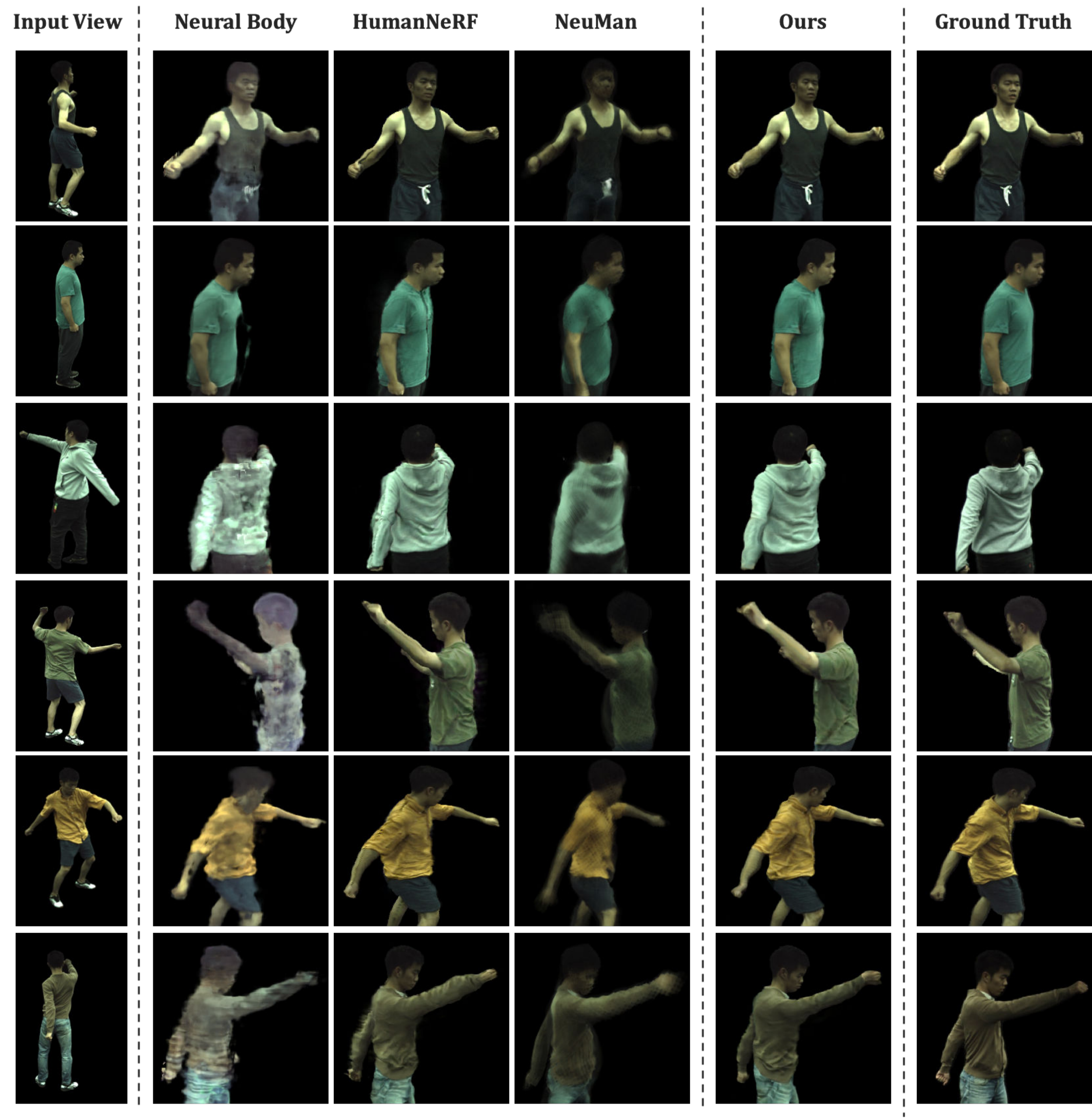}
    \vspace{-4mm}
    \caption{\textbf{Qualitative result of novel view setting in ZJU-MoCap.} We compare the novel view synthesis quality with baseline methods in ZJU-Mocap. Result shows that our method synthesizes more realistic images in novel view.}
    \label{fig:nv_zju}
    \vspace{-5mm}
\end{figure*}

\subsection{Forward Correspondence Search Module}
\label{sec:3.3}
\begin{figure}[h]
    \centering
    \vspace{-3mm}
    \includegraphics[width=0.98\linewidth]{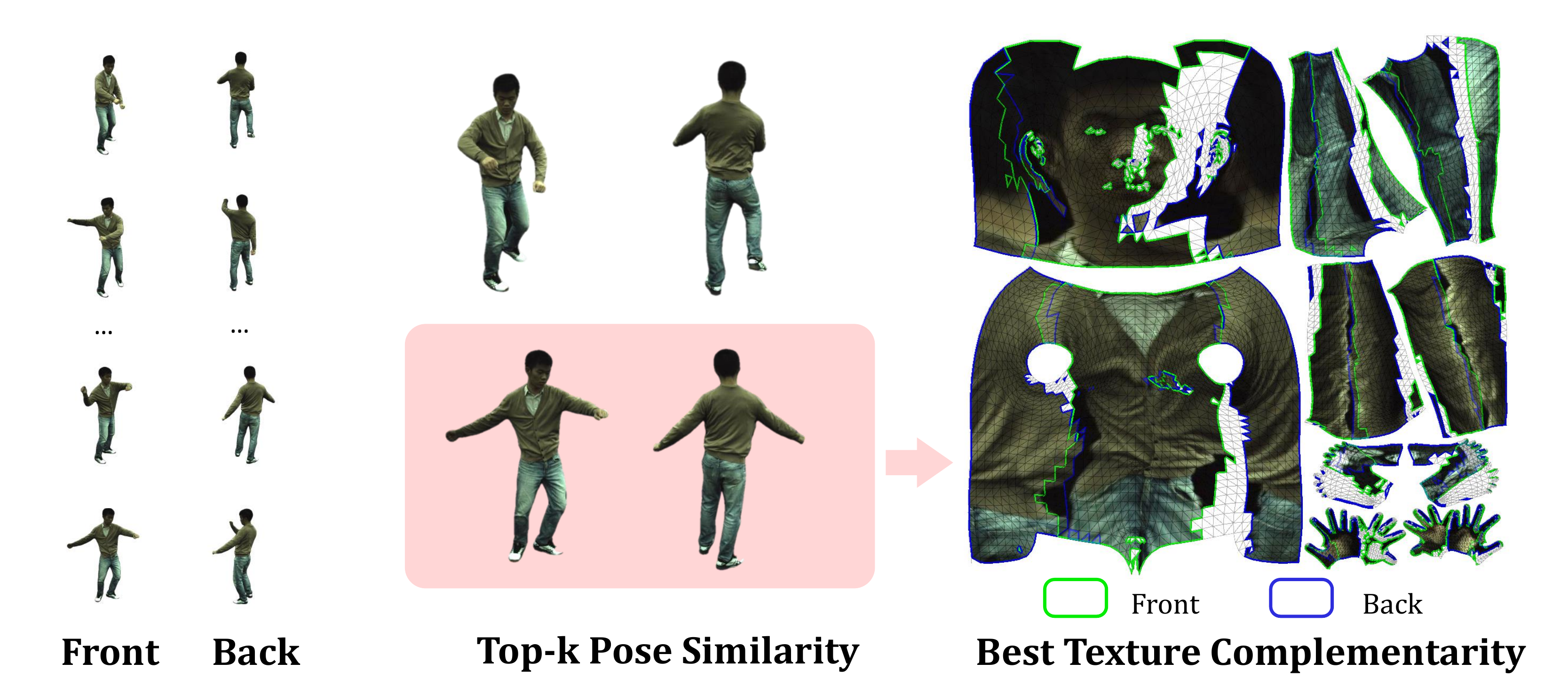}
    \vspace{-4mm}
    \caption{\textbf{The Illustration of Keyframe Selection.} We select the keyframes according to their pose similarity and texture map complementarity.}
    \label{fig:keyframe}
    \vspace{-3mm}
\end{figure}

Previous multi-view works~\cite{wang2021ibrnet,chen2021mvsnerf,zhao2021humannerf} utilize the correspondence features from observed views at the same time stamp to guide the novel view synthesis. As our input images are from monocular video, we don't have multi-view images at the same time. However, we can utilize the forward deformation in Sec.~\ref{sec:3.2} to achieve dynamic deformation and find correspondences in different time stamps in our monocular video. Inspired by the previous works, We design our observation bank consisting of correspondence features to guide the rendering. We propose to build an observation bank by spanning the time and searching the correspondence features in these posed keyframe images from the input monocular video sequence in order to guide the novel view and novel pose synthesis. As illustrated in Fig.~\ref{fig:keyframe}, we first separate the frames into two subsets containing the front and back view images respectively, according to the orientation of the pelvis. Then, we find the k pairs with the closest pose from these two sets. 
 We reconstruct the texture map for the k pairs, and finally select the pair with the highest texture map complementarity. However, as the lack of accurate camera parameters, searching the correspondence in the monocular input setting is a non-trivial problem. To address this, we utilize the advantage of forward deformation described in Sec.~\ref{sec:3.2} to search the correspondence features of point ${x_c}$ in canonical space. Specifically, we first deform ${x_c}$ to ${x_o}^i$ from canonical to correspondence space using Eq.~\ref{eq:forward-deform-formulate}:
\begin{equation}
{\mathbf{x_o}}^i = \
  {D_f}(\mathbf{{x_c}}, \mathbf{p_i}),
\label{eq:correspondence-deform}  
\end{equation}
where ${\mathbf{p_i}}$ denotes the pose in $i$-th frame. Then with the camera parameters of $i$-th frame, we project the point ${x_o}^i$ to the image coordinate:
\begin{equation}
{\mathbf{x_i}} =
  {K_i}{E_i}{\mathbf{{x_o}^i}},
\label{eq:projection}  
\end{equation}
where ${K_i}$, $E_i$ are the intrinsic and extrinsic camera parameters of $i$-th frame respectively. ${x_i}$ is the pixel location in $i$-th frame, which is used to sample the feature ${f_i}$ and color ${c_i}$ of $i$-th image. We follow the U-Net feature extract from IBRNet~\cite{wang2021ibrnet} to extract the image features. We get the feature of $j$-th frame ${f_j}$ and ${c_j}$ in the same way. Then we use a Blend MLP to map these two features to the blend weights:
\begin{equation}
\begin{aligned}
&{\mathbf{F}} =
  {w_i}{({f_i};{c_i})} + {w_j}{({f_j};{c_j})},\\
&{\mathbf{w}} =
  {\rm MLP_{blend}}({({f_i};{c_i})}, {({f_j};{c_j})}),
 \end{aligned}
\label{eq:blend-computation}  
\end{equation}
where ${(;)}$ is the concatenation of two vectors. The blended feature ${\mathbf{F}}$ is used to guide the rendering network.

\begin{figure}[htbp]
    \centering
    \vspace{-1ex}
    \includegraphics[width=0.94\linewidth]{
    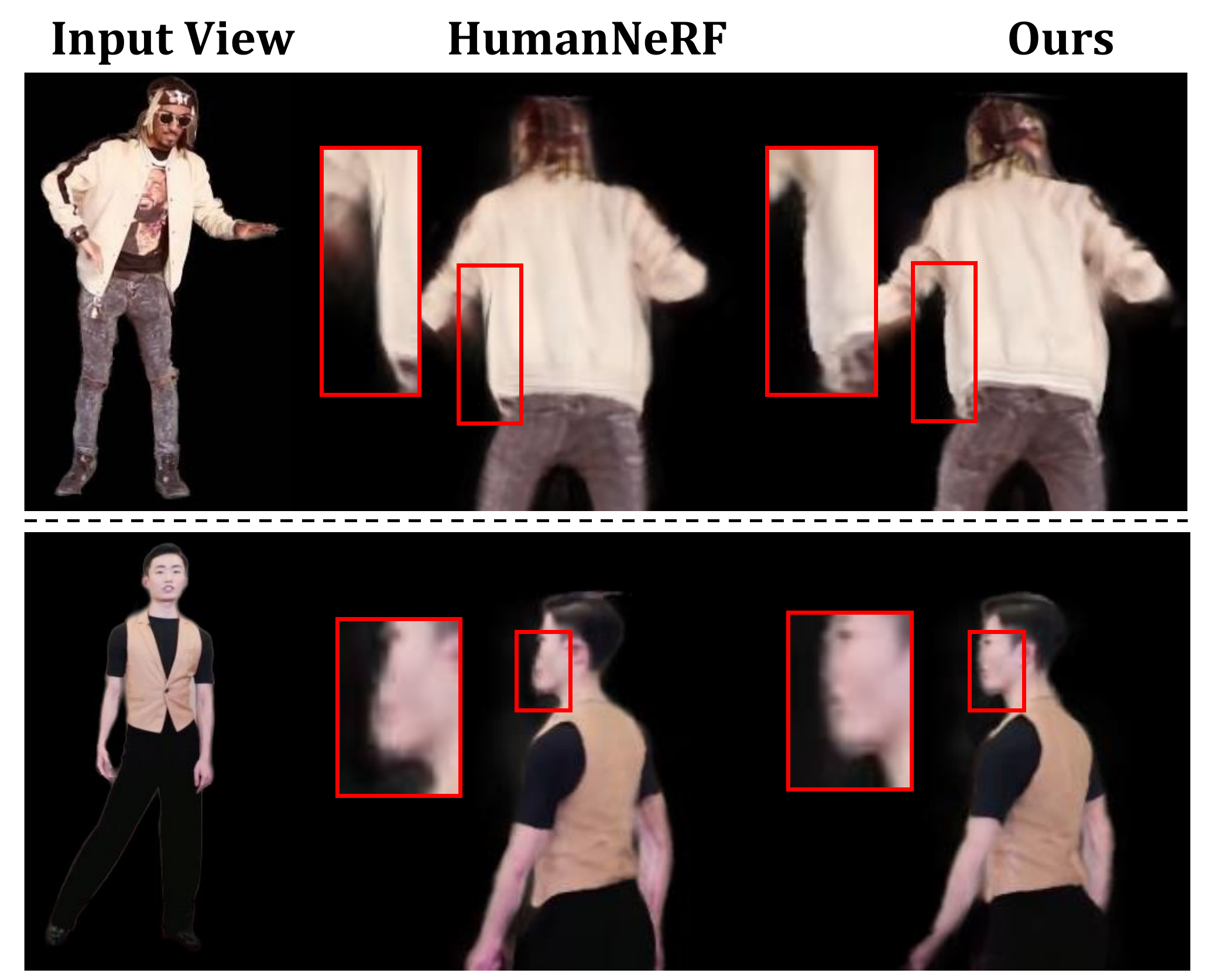}
    \vspace{-2mm}
    \caption{\textbf{Qualitative results on novel view from Internet video.} Top: invisible (\href{https://youtu.be/ANwEiICt7BM}{video source}), Down: Cha-Cha (\href{https://youtu.be/kIEQhedG1Qo}{video source})}
    \label{fig:nv_youtube}
    \vspace{-3mm}
\end{figure}

\subsection{Volume Rendering and Network Training}
\label{sec:3.4}
\noindent
\textbf{Volume rendering with deformation.} We conduct the rendering step in one canonical space. With the guidance of the feature extracted by the Forward Correspondence Search module described in Sec.~\ref{sec:3.3}, our rendering network takes input canonical space point $\mathbf{x_i}$ and feature $\mathbf{F}$, and outputs its color $\mathbf{c}({\mathbf{x_i}})$ and density $\mathbf{\sigma}(\mathbf{x_i})$. We denote our whole rendering network as a mapping ${F_r}$:
\begin{equation}
\mathbf{c}({\mathbf{x_i}}), \mathbf{\sigma}(\mathbf{x_i}) =
  {F_r}(\gamma(\mathbf{x_i}), \mathbf{F}),
\label{eq:rendering}  
\end{equation}
where $\gamma$ is a sinusoidal positional encoding~\cite{mildenhall2020nerf} of ${x_i}$.
We finally render the neural field in canonical space using the volume rendering equation~\cite{max1995optical, mildenhall2020nerf}. The color $C(\mathbf{r})$ of a ray $\mathbf{r}$ with $D$ samples can be written as:
\begin{equation}
\label{eq:volume_rendering}
\begin{aligned}
    & C(\mathbf{r}) = \sum_{i=1}^{D} (\prod_{j=1}^{i-1} (1 - \alpha_j))\alpha_i\mathbf{c}(\mathbf{x_i}), \\
    & \quad \alpha_i = 1 - \exp(-\mathbf{\sigma}(\mathbf{x_i}) \Delta t_i), 
\end{aligned}
\end{equation}
where $\Delta t_i$ is the interval between sample $i$ and $i+1$. 

\noindent
\textbf{Network training.} We optimize our network with $\mathcal{L}_{\text{MSE}}$, $\mathcal{L}_{\text{LPIPS}}$ and $\mathcal{L}_{\text{CONSIS}}$:
\begin{equation}
\label{eq:volume_rendering}
\mathcal{L}_{\text{MSE}} = \sum_{\mathbf{r}\in R}^{}(\left \| C(\mathbf{r}) - \hat{C}(\mathbf{r})  \right \|_2^2)
\end{equation}
where $C(\mathbf{r})$, $\hat{C}(\mathbf{r})$ are rendered color and ground truth color of camera ray $\mathbf{r}$. We also adopt a perceptual loss LPIPS~\cite{zhang2018unreasonable} in our final loss. The consistent loss described in Sec.~\ref{sec:3.2} is used for optimization, with the final loss function as:
\begin{equation}
\label{eq:volume_rendering}
\mathcal{L} = \mathcal{L}_{\text{MSE}} + \lambda \mathcal{L}_{\text{LPIPS}} + \mathcal{L}_{\text{CONSIS}}
\end{equation}
where $\lambda$ is a weight balance coefficient.

\section{Experiments}
\label{sec:experiment}

\subsection{Dataset and Preprocessing}
We use ZJU-MoCap dataset\cite{peng2021neural} and in-the-wild video collected from the Internet to evaluate our method. We follow\cite{weng2022humannerf} to select the same 6 subjects for our evaluation. As ZJU-MoCap is carefully collected in an indoor environment, we directly use the annotations in the dataset. For the wild video we gathered from the Internet, we used PARE\cite{kocabas2021pare} to estimate the camera matrix and body pose, and RVM\cite{lin2022robust} to get the segmentation mask. We select keyframes as illustrated in Fig.~\ref{fig:keyframe} and Sec.~\ref{sec:3.3}.

\subsection{Experimental Settings}
\noindent
\textbf{Comparation baselines.} We compare our method with (1) NeuralBody~\cite{peng2021neural}, which uses structured latent code to represent the human body; (2) HumanNeRF~\cite{weng2022humannerf}, which achieves state-of-the-art image synthesis performance of digital human by learning a motion field mapping network from monocular video. (3) NeuMan~\cite{jiang2022neuman}, who reconstructs the scene and human using two different NeRFs, we only compare with NeuMan's human NeRF without background. 

\noindent
\textbf{Implementation details.}
We adopt a U-Net to extract image features from keyframes similar to ~\cite{wang2021ibrnet}, and the deform error thread $\theta=0.05$. During training, the learning rate of the Shared Bidirectional Deformation module is set to $5\times10^{-6}$ and $5\times10^{-5}$ for the rest. The overall framework trains on a single V100 GPU for 70 hours.

\begin{table}[t]
\centering
\begin{tabular}{c|c|c|c}
\Xhline{1.2pt}
 & PSNR $\uparrow$ & SSIM $\uparrow$ & LPIPS* $\downarrow$ \\
\Xhline{1.2pt}
Neural Body \cite{peng2021neural}&  28.72&  0.9611& 52.25 \\
HumanNeRF \cite{weng2022humannerf}& 29.61 & 0.9625 & 38.45 \\
NeuMan \cite{jiang2022neuman}& 28.96 & 0.9479 & 60.74 \\
\Xhline{1.2pt}
\textbf{Ours} & \textbf{30.26} & \textbf{0.9692} & \textbf{30.92} \\
\Xhline{1.2pt}
\end{tabular}
\vspace{-3px}
\caption{\textbf{Novel view synthesis quantitative comparison on ZJU-MoCap dataset.} We compute averages over 6 sequences. For results of each subject please refer to the appendix. Note that in all table results, LPIPS* = LPIPS $\times 10^3$.}
\vspace{-4ex}
\label{table:nv_zju_number_vs_nb_hn}
\end{table}

\subsection{Quantitative Evaluation}

\noindent
\textbf{Evaluation metrics.} We use PSNR and SSIM~\cite{wang2004image} to evaluate the generated image quality. As PSNR prefers smooth results but may with bad visual quality~\cite{zhang2018unreasonable, weng2022humannerf}, we additionally adopt LPIPS~\cite{zhang2018unreasonable} to measure the perceived distance between the synthesis image and ground truth image.

\noindent
\textbf{Comparison settings.} For comparison of the performance of novel view synthesis and animation, we follow~\cite{weng2022humannerf} to only use the camera 1 data in ZJU-MoCap for training. We further split the frames in camera 1 as seen-pose Set A and unseen-pose Set B in 4:1 ratio. We compare our method with baselines in two settings. 1) The novel view setting in Tab.~\ref{table:nv_zju_number_vs_nb_hn}, where we follow\cite{weng2022humannerf} to use frames in Set A as training and the synchronous frames in the other 22 cameras as evaluation. 2) The novel pose setting in Tab.~\ref{table:np_zju_number_vs_nb_hn}, where we use frames in Set A as training, the frames in Set B, and its synchronous frames in other 22 cameras as evaluation. 

\noindent
\textbf{Evaluation results.} The results of the novel view setting
and novel pose setting are shown in Tab.~\ref{table:nv_zju_number_vs_nb_hn} and Tab. \ref{table:np_zju_number_vs_nb_hn}, respectively. We can see that our MonoHuman framework outperforms existing methods in most metrics in both settings.
With the help of correspondence features and the consistent constraint of the deformation field, we can synthesize more realistic results. Though the PSNR metric of NeuralBody is not bad, they synthesize poor visual quality images in both novel view and novel pose (as PSNR prefers smooth results\cite{zhang2018unreasonable}, but it can be verified by LPIPS). Our improvement in PSNR over HumanNeRF\cite{weng2022humannerf} is not significant, and slightly lower in subject 387 when testing in the novel pose.  But in LIPIS, our method has about 19.6\% and 9.46\% improvement in novel view and novel pose setting respectively. Note that novel pose synthesis is more challenging than novel view synthesis, but in some subjects, the evaluation metrics behave the opposite because some poses are repeated in the data. So we conduct extract qualitative comparison in out of distribution pose. NeuMan~\cite{jiang2022neuman} only updates the SMPL parameter during training, and models the deformation as the closest point's LBS weight, which leads to poor performance in novel views.

\renewcommand{\arraystretch}{1.2}
\begin{table}[t]
\centering
\begin{tabular}{c|c|c|c}
\Xhline{1.2pt}
 & PSNR $\uparrow$ & SSIM $\uparrow$ & LPIPS* $\downarrow$ \\
\Xhline{1.2pt}
Neural Body \cite{peng2021neural}&  28.54&  0.9604& 53.91 \\
HumanNeRF \cite{weng2022humannerf}& 29.74 & 0.9655 & 34.79 \\
NeuMan \cite{jiang2022neuman}& 28.75 & 0.9406 & 62.35 \\
\Xhline{1.2pt}
\textbf{Ours}  & \textbf{30.05} & \textbf{0.9684} & \textbf{31.51} \\
\Xhline{1.2pt}
\end{tabular}
\vspace{-2px}
\caption{\textbf{Novel pose synthesis quantitative comparison on ZJU-MoCap dataset.} We compute averages over 6 sequences. For results of each subject please refer to the appendix.}
\label{table:np_zju_number_vs_nb_hn}
\vspace{-4ex}
\end{table}

\begin{figure*}[t!]
    \centering
    \vspace{-4mm}
    \includegraphics[width=0.96\linewidth]{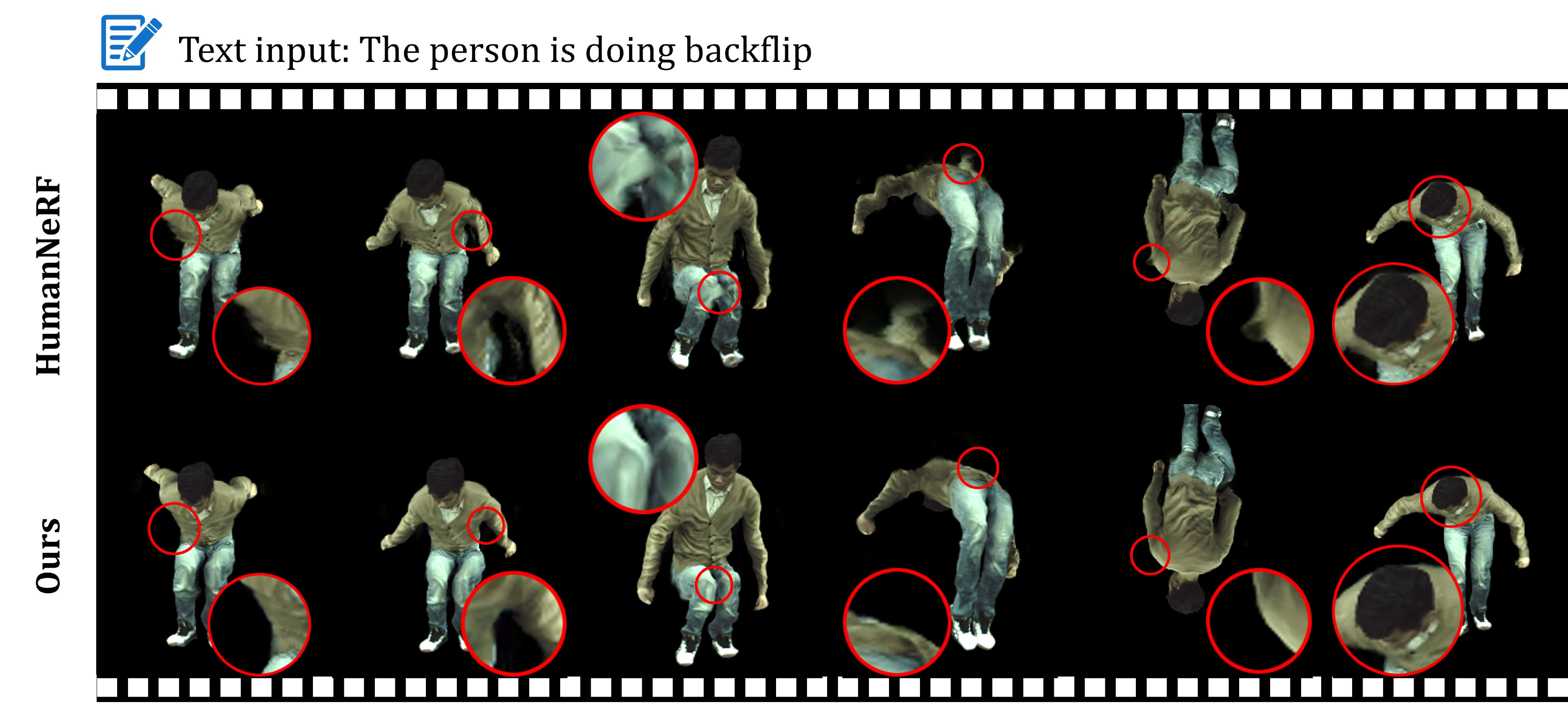}
    \vspace{-2mm}
    \caption{\textbf{Qualitative results on challenge poses generated by MDM~\cite{tevet2022human} through text input.} We evaluate our method driven by challenge pose sequence generated by MDM model.}
    \label{fig:mdm_poses}
\end{figure*}

\subsection{Qualitative Evaluation}
To compare the generated results, we visualize the novel view synthesis results in ZJU-MoCap dataset of NeuralBody, HumanNeRF and our method in Fig.\ref{fig:nv_zju}. NeuralBody tends to generate vague images with large noise in novel view. Both HumanNeRF and our method synthesize realistic images in novel view but predict artifacts in some detail areas. We also show the novel view synthesis results in videos collected from the Internet in Fig.\ref{fig:nv_youtube}. For extreme views, HumanNeRF tends to generate artifacts in clothes and face areas. while our methods can retain these detail due to the correct deform and help with guided features.
\par As the motions in ZJU-MoCap and self-collected videos are simple and usually repetitive, we need to find a way to evaluate the animation ability in some complex motions. To do this, we use the MDM~\cite{tevet2022human}, which can generate complex motions corresponding to the text input. We use such a model to generate challenge motions like back-flip and martial pose sequences to animate the avatar reconstructed by HumanNerf and our method. As shown in Fig.~\ref{fig:mdm_poses}, HumanNeRF predicts multiple artifacts in some extreme poses like squatting  and jumping high. It also fails to deform the arms correctly in punch motions and back-flip poses in the air. With the help of the Shared Bidirectional Deformation Module, we proposed, we can handle the deformation in such challenging poses and generate more realistic results.

\begin{table}[ht]
\centering
\begin{tabular}{c|c|c|c}
\Xhline{1.2pt}
 & PSNR $\uparrow$ & SSIM $\uparrow$ & LPIPS* $\downarrow$ \\
\Xhline{1.2pt}
Neural Body \cite{peng2021neural}&  28.72&  0.9611& 52.25 \\
HumanNeRF \cite{weng2022humannerf}& 29.61 & 0.9625 & 38.45 \\
NeuMan \cite{jiang2022neuman}& 28.96 & 0.9479 & 60.74 \\
\hline
Ours (w/o $\mathcal{L}_{\text{CONSIS}}$) &30.17  &0.9689  &31.76   \\
Ours (w/o feat) &30.16  &0.9689  &31.39   \\
\Xhline{1.2pt}
\textbf{Ours (full model)} & \textbf{30.26} & \textbf{0.9692} & \textbf{30.92} \\
\Xhline{1.2pt}
\end{tabular}
\vspace{-3px}
\caption{\textbf{Ablation study on ZJU-MoCap in novel view setting.} We compute averages over 6 sequences. LPIPS* = LPIPS $\times 10^3$.}
\vspace{-3px}
\label{table:ablation_zju_nv}
\end{table}

\begin{table}[t]
\centering
\begin{tabular}{c|c|c|c}
\Xhline{1.2pt}
 & PSNR $\uparrow$ & SSIM $\uparrow$ & LPIPS* $\downarrow$ \\
\Xhline{1.2pt}
Neural Body \cite{peng2021neural}&  28.54&  0.9604& 53.91 \\
HumanNeRF \cite{weng2022humannerf}& 29.74 & 0.9655 & 34.79 \\
NeuMan \cite{jiang2022neuman}& 28.75 & 0.9406 & 62.35 \\
\hline
Ours (w/o $\mathcal{L}_{\text{CONSIS}}$) & 29.86 & 0.9679 & 32.39  \\
Ours (w/o feat) & 29.88 & 0.9681 & 31.80  \\
\Xhline{1.2pt}
\textbf{Ours (full model)} & \textbf{30.05} & \textbf{0.9684} & \textbf{31.51} \\
\Xhline{1.2pt}
\end{tabular}
\vspace{-2px}
\caption{\textbf{Ablation study on ZJU-MoCap in novel pose setting.} We compute averages over 6 sequences. LPIPS* = LPIPS $\times 10^3$.}
\label{table:ablation_zju_np}
\end{table}

\subsection{Ablation Study}
We conduct ablation studies for the two key modules in our model, the Shared Bidirectional Deformation module and the Forward Correspondence Search module in ZJU-MoCap dataset. To further explore the function of these two modules, we experiment in novel view and novel pose settings separately. The result is shown in Tab.~\ref{table:ablation_zju_nv} and Tab.~\ref{table:ablation_zju_np} respectively. As shown in Tab.~\ref{table:ablation_zju_nv}, these two modules contribute roughly the same to the result. Because the input poses in the novel view setting are all seen in training data, so even without the consistency constraints, the deformation is mostly correct. But when it comes to the novel poses, LPIPS is higher as the deformation is more difficult in the novel view setting. Furthermore, the incorrect deformation will influence the correspondence features we searched by our FCS module, which makes the generated result worse.\\
\indent The influence without the Forward Correspondence Search module is nearly the same in both novel view and novel pose settings, as the guidance of correspondence feature is equally important in both novel view and novel pose settings. However, the performance of our FCS module relies on the Shared Bidirectional Deformation module, and the result of Tab.~\ref{table:ablation_zju_np} verifies this property. In comparison to other baselines, these two modules we proposed are helpful to the generated results. We furthermore visualize the effeteness of our consistent loss, please refer to the appendix for more details.

\section{Discussion}
\label{sec:conclusion}
 We propose MonoHuman, which robustly renders view-consistent avatars at novel poses of high fidelity. We propose a novel Shared Bidirectional Deformation module to handle out-of-distribution novel pose deformation. Furthermore, a Forward Correspondence Search module which queries the correspondence feature to guide the rendering network is used to help generate photo-realistic results. To evaluate our approach, we generate challenging poses by Human Motion Diffusion Model~\cite{tevet2022human} to animate our avatar. The result shows that our model can generate high-fidelity images even under challenging novel pose settings.
 
\noindent
\textbf{Limitations and Future work.}
Our methods can create an animatable avatar with only a monocular video. However, our method still has a few limitations: 1) Our synthesis result relies on the accuracy of the pose and mask annotations. It would be interesting to explore how to make pose and mask correction during training. 2) Our method is trained in a case-specific manner. Studying how to generalize different people with monocular videos will be valuable.

{\small
\bibliographystyle{ieee_fullname}
\bibliography{egbib}
}

\newpage

\appendix

\noindent
\textbf{\LARGE Appendix}

\setcounter{table}{0}
\renewcommand{\thetable}{A\arabic{table}}
\setcounter{figure}{0}
\renewcommand{\thefigure}{A\arabic{figure}}

\section{Network Architecture}
\subsection{Blend MLP}
\begin{figure}[htbp]
    \centering
    \vspace{-1ex}
    \includegraphics[width=0.8\linewidth]{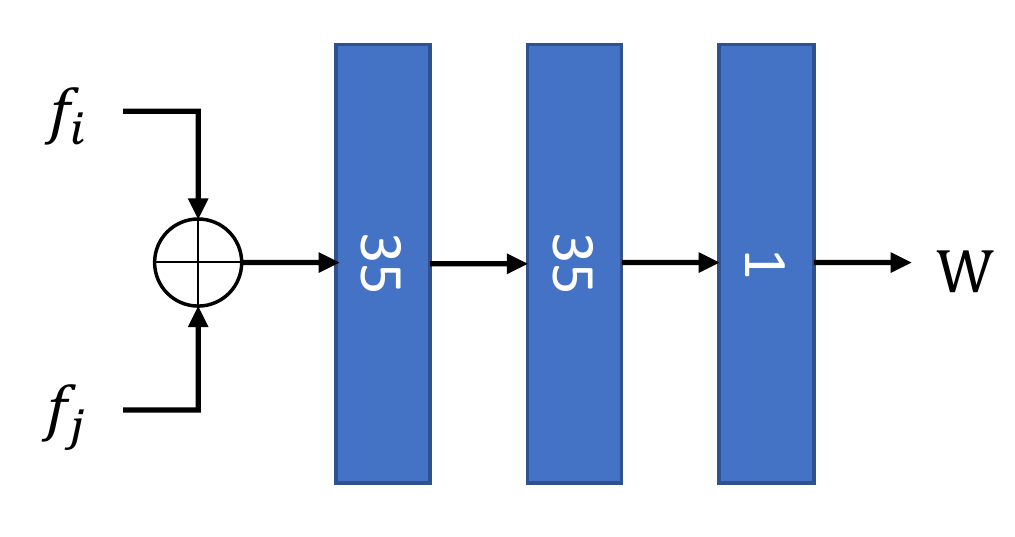}
    \vspace{-2mm}
    \caption{\textbf{Visualization of Blend MLP.} We use 3 layers MLP to calculate the belnd weights of correspondence features. It takes the concatenation of two correspondence features $f_i$ and $f_j$ as input, and output the final blend weight $\mathbf{W}$ of these two features.}
    \label{fig:BlendMLP}
    \vspace{-3mm}
\end{figure}

\subsection{Rendering Network}
\begin{figure}[htbp]
    \centering
    \vspace{-1ex}
    \includegraphics[width=0.94\linewidth]{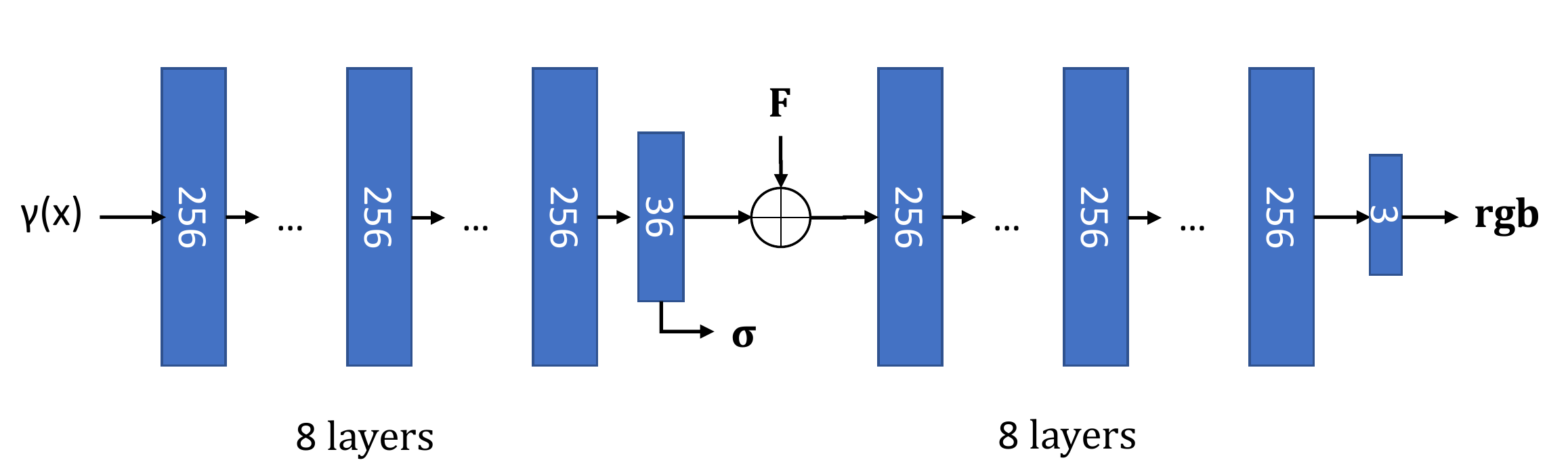}
    \vspace{-2mm}
    \caption{\textbf{Visualization of Rendering Network.} We use total 16 layers MLP to calculate the final color rgb and density $\sigma$. The first 8 layers of MLP take $\gamma(x)$ which means the positional encoding of point $x$ as input, and output a 36 dimensional vector. One dimension of first 8 layers MLP output density value $\sigma$, and the remaining dimension is concatenated with the blended feature $F$. The second 8 layers MLP take the concatenation as input, and output the final rgb color value.}
    \label{fig:rendering}
    \vspace{-3mm}
\end{figure}

\section{Ablation Study}
\subsection{Ablation Study on Shared Bidirectional Deformation Module}
\begin{figure}[htbp]
    \centering
    \vspace{-3mm}
    \includegraphics[width=0.94\linewidth]{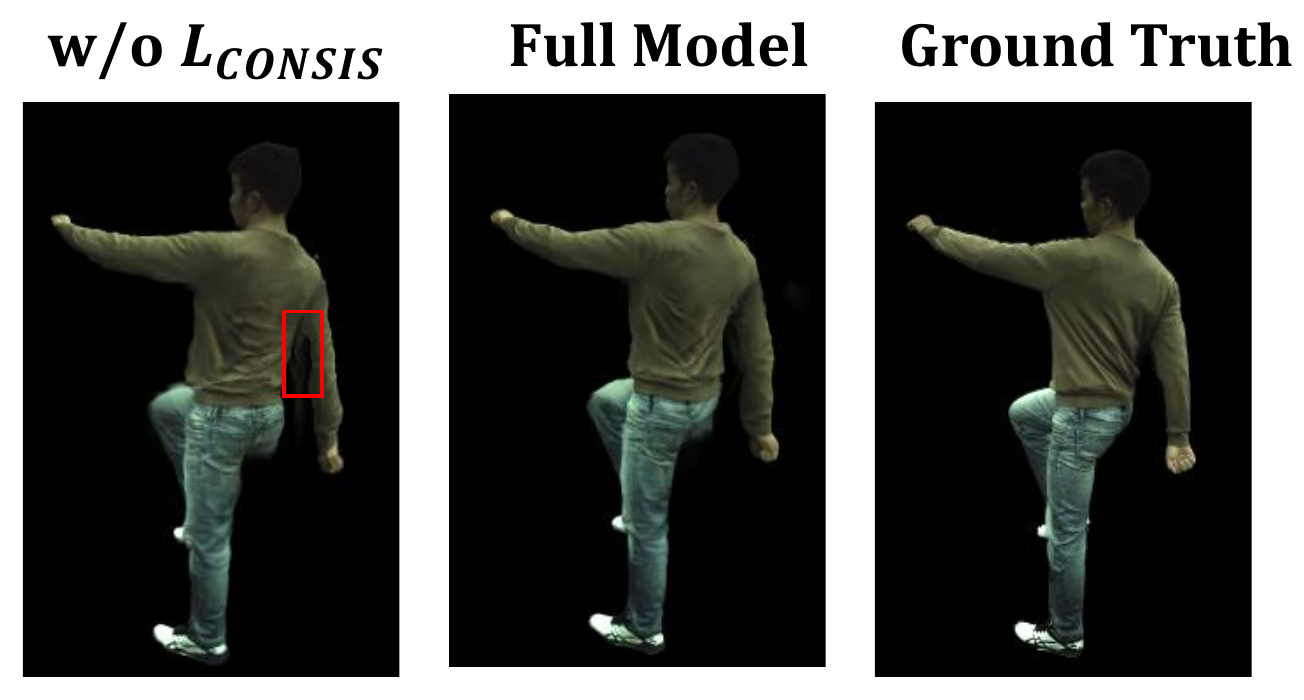}
    \vspace{-2mm}
    \caption{\textbf{Ablation of $\mathcal{L}_{\text{CONSIS}}$.} We compare the qualitative result without consistent loss.}
    \label{fig:consis}
    \vspace{-3mm}
\end{figure}
Fig.~\ref{fig:consis} illustrates that the Shared Bidirectional Deformation Module with consistent loss we proposed help produce more accurate deformation in regions like arms. Without this loss, the deformation in arms area tends to be bending and produce obvious artifacts.

\begin{table*}[htbp]
\begin{tabular}{|l|lll|lll|lll|}
\hline
& \multicolumn{3}{l|}{PSNR}                                                     & \multicolumn{3}{l|}{SSIM}                                                     & \multicolumn{3}{l|}{LPIPS}                                                    \\ \hline
frame\_nums & \multicolumn{1}{l|}{w/o $\mathcal{L}_{\text{CONSIS}}$} & \multicolumn{1}{l|}{w/o feat} & full & \multicolumn{1}{l|}{w/o $\mathcal{L}_{\text{CONSIS}}$} & \multicolumn{1}{l|}{w/o feat} & full & \multicolumn{1}{l|}{w/o $\mathcal{L}_{\text{CONSIS}}$} & \multicolumn{1}{l|}{w/o feat} & full \\ \hline
380         & \multicolumn{1}{l|}{27.87}       & \multicolumn{1}{l|}{27.84}    & \textbf{27.89}      & \multicolumn{1}{l|}{0.9387}      & \multicolumn{1}{l|}{0.9389}   & \textbf{0.9389}     & \multicolumn{1}{l|}{55.85}       & \multicolumn{1}{l|}{\textbf{54.91}}    & 55.28      \\ \hline
76          & \multicolumn{1}{l|}{27.44}       & \multicolumn{1}{l|}{27.71}    & \textbf{27.77}      & \multicolumn{1}{l|}{0.9348}      & \multicolumn{1}{l|}{0.9368}   & \textbf{0.9390}     & \multicolumn{1}{l|}{63.16}       & \multicolumn{1}{l|}{60.94}    & \textbf{56.40}      \\ \hline
38          & \multicolumn{1}{l|}{27.65}       & \multicolumn{1}{l|}{27.71}    & \textbf{27.88}      & \multicolumn{1}{l|}{0.9366}      & \multicolumn{1}{l|}{0.9358}   & \textbf{0.9390}     & \multicolumn{1}{l|}{61.94}       & \multicolumn{1}{l|}{62.32}    & \textbf{56.40}      \\ \hline
19          & \multicolumn{1}{l|}{27.27}       & \multicolumn{1}{l|}{27.35}    & \textbf{27.52}      & \multicolumn{1}{l|}{0.9341}      & \multicolumn{1}{l|}{0.9359}   & \textbf{0.9361}     & \multicolumn{1}{l|}{67.56}       & \multicolumn{1}{l|}{62.82}    & \textbf{62.56}      \\ \hline
\end{tabular}
\caption{\textbf{Ablation study on Sequence Length.} We compare novel view synthesis results in different training frame length. LPIPS* = LPIPS $\times 10^3$.}
\vspace{-2mm}
\label{table:ablation_seq}

\end{table*}

\begin{figure}[htbp]
    \centering
    \vspace{-4mm}
    \includegraphics[width=0.96\linewidth]{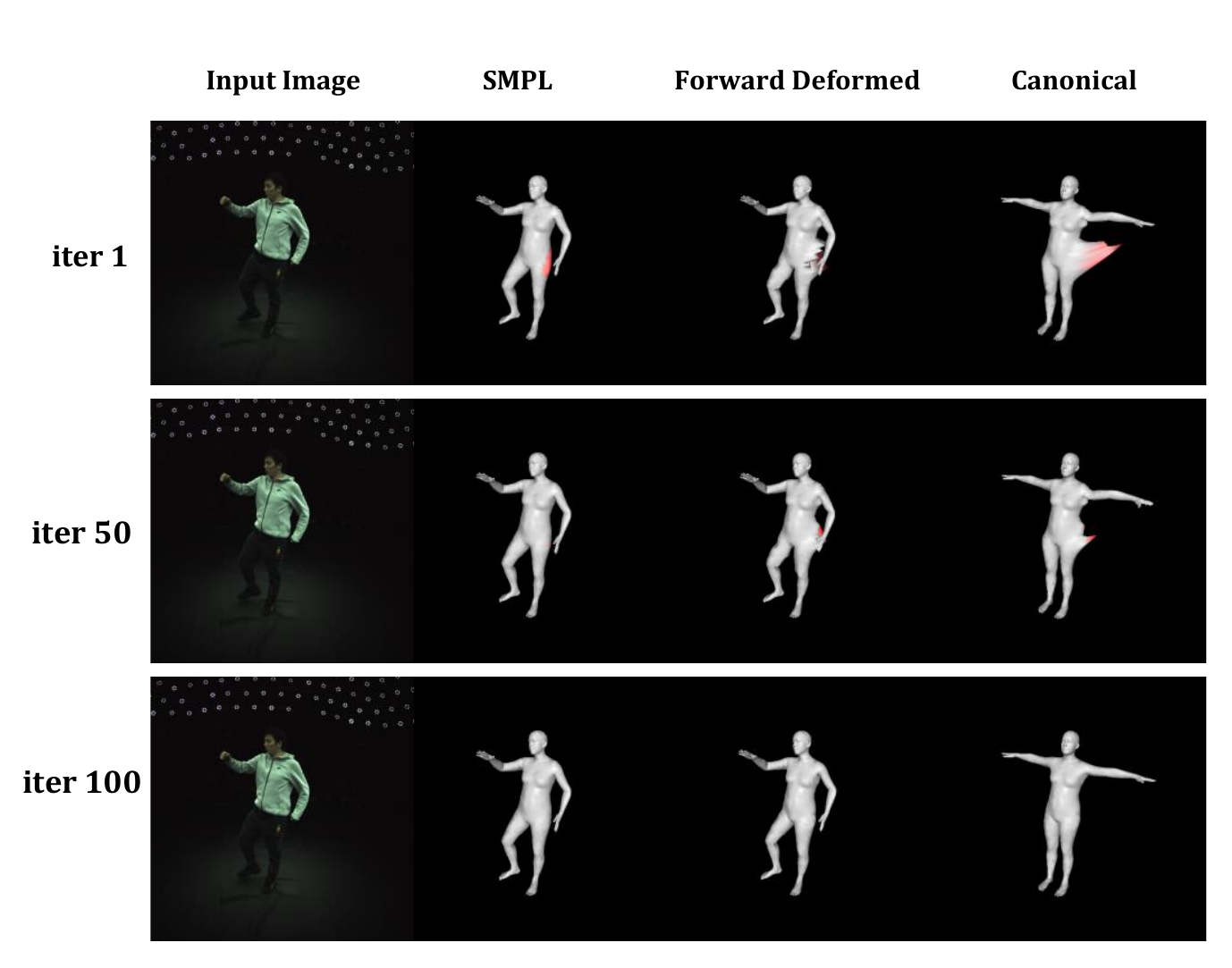}
    \vspace{-2mm}
    \caption{\textbf{Visualization of $\mathcal{L}_{\text{CONSIS}}$.} We optimize the $\mathcal{L}_{\text{CONSIS}}$ separately and visualize its effectiveness.}
    \label{fig:vis_consis}
    \vspace{-3mm}
\end{figure}

\begin{figure}[htbp]
    \centering
    \vspace{-5mm}
    \includegraphics[width=0.97\linewidth]{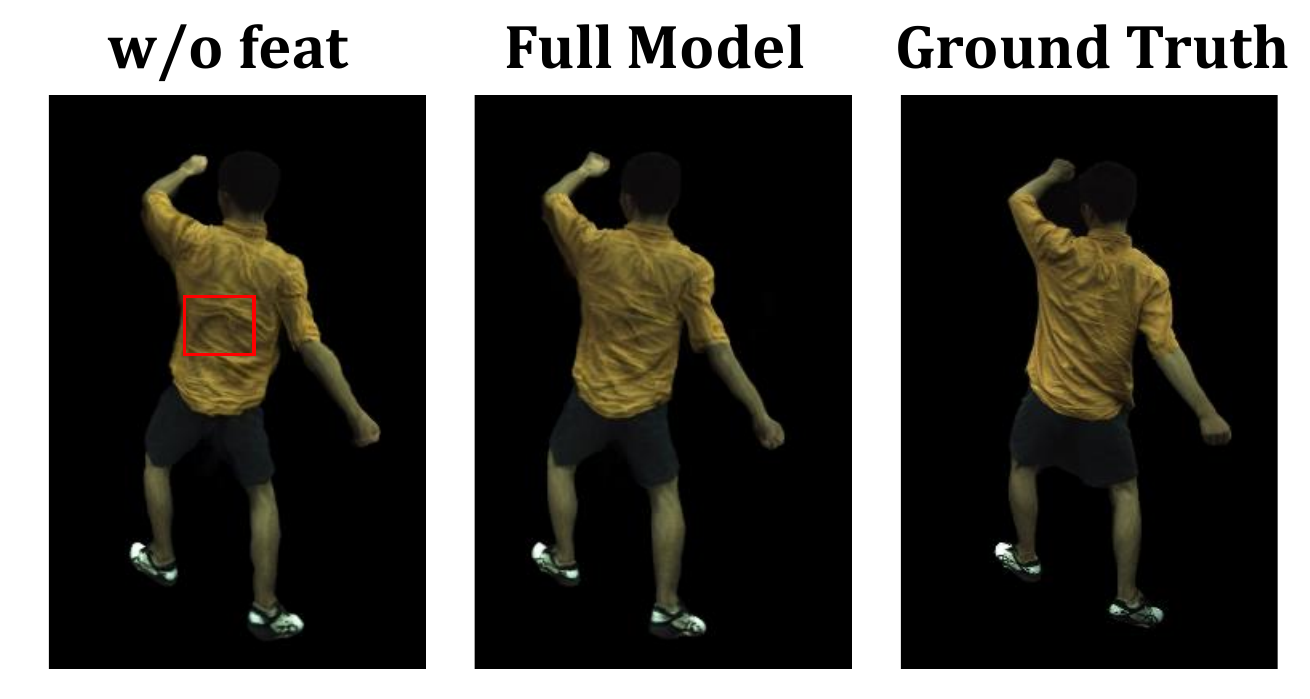}
    \vspace{-2mm}
    \caption{\textbf{Qualitative ablation of correspondence features.} We compare qualitative results with/without correspondence features.}
    \label{fig:feat}
    \vspace{-3mm}
\end{figure}

In Fig.~\ref{fig:vis_consis}, we use SMPL vertex as input of Shared Bidirectional Deformation Module and optimize it with a learning rate of $5 \times 10^{-6}$ separately. Points calculated by the consistent loss which value $>= 0.05$ is colored red. The second column is the projection of the input image's SMPL vertex. Canonical Vertex is the points deformed from SMPL vertex using backward deform of Shared Bidirectional Deformation Module. Forward Deformed Vertex means the points forward deformed from canonical vertex. The red points illustrate that our consistent loss can detect the points that are deformed incorrectly. We further visualize the results of iteration1, 50 and 100. It shows that the red points reduce gradually, which means that the consistent loss corrects and regularizes the shared deformation weight.

\subsection{Ablation Study on Forward Correspondence Search Module}

Fig.~\ref{fig:feat} shows that the correspondence features produced by the Forward Correspondence Search module we proposed help produce more accurate color and texture in cloth regions. Without these features, the synthesis result tends to produce an unnatural texture in the cloth.

\subsection{Ablation Study on Sequence Length}
In order to explore our method's performance in different training sequence lengths, we evaluate novel view synthesis results in different sample rates. The result is shown in Table ~\ref{table:ablation_seq}. We use subject 394 in ZJU-MoCap as testing and sample in rates of 1, 5, 10, and 20, and the number of frames is 380, 76, 38, 19 respectively. We follow HumanNeRF~\cite{weng2022humannerf} to evaluate in 22 cameras not seen in training in 30 sample rates. And follow NeuralBody~\cite{peng2021neural} to evaluate the subject in a 3d bounding box to avoid getting an inflated PSNR value. The result shows that the relationship between sequence length and generated quality is not linear. When under small training frame numbers like 19 frames, our module helps to retain the more realistic result.

\begin{table*}[t]
\centering
\resizebox{1\textwidth}{!}{
\begin{tabular}{|c || c | c | c || c | c | c || c | c | c|}
\hline
\multirow{2}{*}{} &  \multicolumn{3}{c||}{Subject \textbf{377}} & \multicolumn{3}{c||}{Subject \textbf{386}} & \multicolumn{3}{c|}{Subject \textbf{387}} \\ 
\cline{2-10}
 & PSNR $\uparrow$ & SSIM $\uparrow$ & LPIPS* $\downarrow$ & PSNR $\uparrow$ & SSIM $\uparrow$ & LPIPS* $\downarrow $ & PSNR $\uparrow$ & SSIM $\uparrow$ & LPIPS* $\downarrow$ \\ 
\hline
Neural Body~\cite{peng2021neural} & 29.29 & 0.9693 & 39.40 
& 30.71 & 0.9661 & 45.89 
& 26.36 & 0.9520 & 62.21 \\
\hline
HumanNeRF~\cite{weng2022humannerf} & 29.91 & 0.9755 & 23.87 
& 32.62 & 0.9672 & 39.36 
& \textbf{28.01} & 0.9634 & 35.27 \\
\hline
Ours & \textbf{30.77}  & \textbf{0.9787} & \textbf{21.67}
& \textbf{32.97}  & \textbf{0.9733} & \textbf{32.73}
& 27.93 & \textbf{0.9633} & \textbf{33.45} \\
\hline
\hline
\multirow{2}{*}{} &  \multicolumn{3}{c||}{Subject \textbf{392}} & \multicolumn{3}{c||}{Subject \textbf{393}} & \multicolumn{3}{c|}{Subject \textbf{394}} \\ 
\cline{2-10}
 & PSNR $\uparrow$ & SSIM $\uparrow$ & LPIPS* $\downarrow$ & PSNR $\uparrow$ & SSIM $\uparrow$ & LPIPS* $\downarrow $ & PSNR $\uparrow$ & SSIM $\uparrow$ & LPIPS* $\downarrow$ \\ 
\hline
Neural Body~\cite{peng2021neural} & 28.97 & 0.9615 & 57.03
& 27.82 & 0.9577 & 59.24
& 28.09 & 0.9557 & 59.66\\
\hline
HumanNeRF~\cite{weng2022humannerf} & 30.95 & 0.9687 & 34.23 
& 28.43 & 0.9609 & 36.26 
& 28.52 & 0.9573 & 39.75 \\
\hline
Ours & \textbf{31.24} & \textbf{0.9715} & \textbf{31.04}
& \textbf{28.46} & \textbf{0.9622} & \textbf{34.24} 
& \textbf{28.94} & \textbf{0.9612} & \textbf{35.90}\\
\hline
\end{tabular}
}
\caption{\textbf{Novel view synthesis quantitative comparison on ZJU-MoCap dataset.} We show the results of each subject. LPIPS* = LPIPS $\times 10^3$.}
\label{table:detail_np}
\vspace{-1px}
\end{table*}

\begin{table*}[!h]
\resizebox{\textwidth}{!}{
\centering
\begin{tabular}{|c || c | c | c || c | c | c || c | c | c|}
\hline
\multirow{2}{*}{} &  \multicolumn{3}{c||}{Subject \textbf{377}} & \multicolumn{3}{c||}{Subject \textbf{386}} & \multicolumn{3}{c|}{Subject \textbf{387}} \\ 
\cline{2-10}
 & PSNR $\uparrow$ & SSIM $\uparrow$ & LPIPS* $\downarrow$ & PSNR $\uparrow$ & SSIM $\uparrow$ & LPIPS* $\downarrow $ & PSNR $\uparrow$ & SSIM $\uparrow$ & LPIPS* $\downarrow$ \\ 
\hline
Neural Body~\cite{peng2021neural} & 29.08 & 0.9679 & 41.17 
& 29.76 & 0.9647 & 46.96 
& 26.84 & 0.9535 & 60.82 \\
\hline
HumanNeRF~\cite{weng2022humannerf}& 29.79 & 0.9714 & 28.49 
& 32.10 & 0.9642 & 41.84 
& 28.11 & 0.9625 & 37.46 \\
\hline
Ours & \textbf{30.46}  & \textbf{0.9781} & \textbf{20.91} 
&  \textbf{32.99} & \textbf{0.9756} & \textbf{30..97} 
& \textbf{28.40} & \textbf{0.9639} & \textbf{35.06} \\
\hline
\hline
\multirow{2}{*}{} &  \multicolumn{3}{c||}{Subject \textbf{392}} & \multicolumn{3}{c||}{Subject \textbf{393}} & \multicolumn{3}{c|}{Subject \textbf{394}} \\ 
\cline{2-10}
 & PSNR $\uparrow$ & SSIM $\uparrow$ & LPIPS* $\downarrow$ & PSNR $\uparrow$ & SSIM $\uparrow$ & LPIPS* $\downarrow $ & PSNR $\uparrow$ & SSIM $\uparrow$ & LPIPS* $\downarrow$ \\ 
\hline
Neural Body~\cite{peng2021neural} & 29.49 & 0.9640 & 51.06 
& 28.50 & 0.9591 & 57.072
& 28.65 & 0.9572 & 55.78\\
\hline
HumanNeRF~\cite{weng2022humannerf} & 30.20 & 0.9633 & 40.06 
& 28.16 & 0.9577 & 40.85
& 29.28 & 0.9557 & 41.97 \\
\hline
Ours & \textbf{30.98} & \textbf{0.9711} & \textbf{30.80}
& \textbf{28.54} & \textbf{0.9620} & \textbf{34.97}
& \textbf{30.21} & \textbf{0.9642} & \textbf{32.80}\\
\hline
\end{tabular}
}
\caption{\textbf{Novel pose synthesis quantitative comparison on ZJU-MoCap dataset.} We show the results of each subject in the table. LPIPS* = LPIPS $\times 10^3$.}
\label{table:detail_nv}
\vspace{-14px}
\end{table*}

\section{Evaluation details}
In order to evaluate novel view and novel pose synthesis, we split the frames in camera 1 in a 4:1 ratio as Set A and B. We only use Set A for training. For novel view evaluation, we sample the synchronous video frames of Set A for all unseen 22 cameras at the rate of 30. For novel pose evaluation, we sample at the same rate for the synchronous frames of Set B for all cameras. In general, the evaluation frames for the novel view would be 242 frames and 184 frames for the novel pose setting.

\section{Condition on view direction}
\begin{figure}[h]
    \centering
    \vspace{-3mm}
    \includegraphics[width=0.98\linewidth]{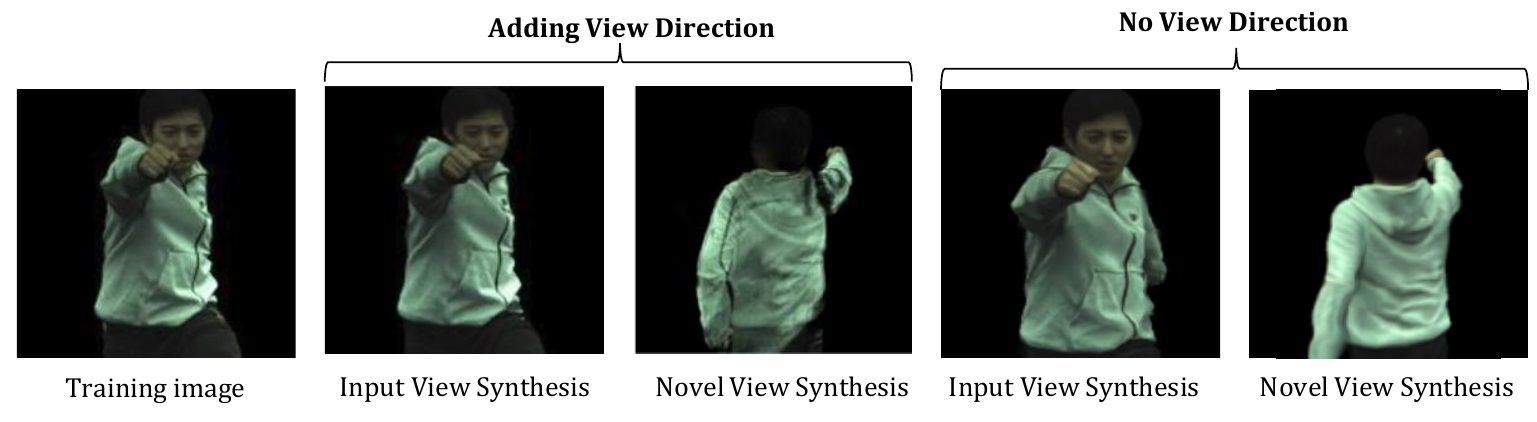}
    \caption{\textbf{Comparison of adding view direction.} We compare the results of conditioning blend weights on view direction and no condition.}
    \label{fig:view_direction}
    \vspace{-5mm}
\end{figure}

We have tried to condition our feature blend weights on view direction when designing our model, but we found overfitting problems that were also found by works like StyleSDF~\cite{or2022stylesdf} and StyleNeRF~\cite{gu2021stylenerf}. As shown in Fig.~\ref{fig:view_direction}, conditioning weights on view direction can help to overfit in the training frames and synthesize realistic results even in some highly reflective areas, but generate lots of artifacts when synthesizing novel view images. We find conditioning blend weights on view direction weakens the generalization ability to the novel view.

\section{More results}
\begin{figure*}[t!]
    \vspace{-3mm}
    \centering
    \includegraphics[width=\linewidth]{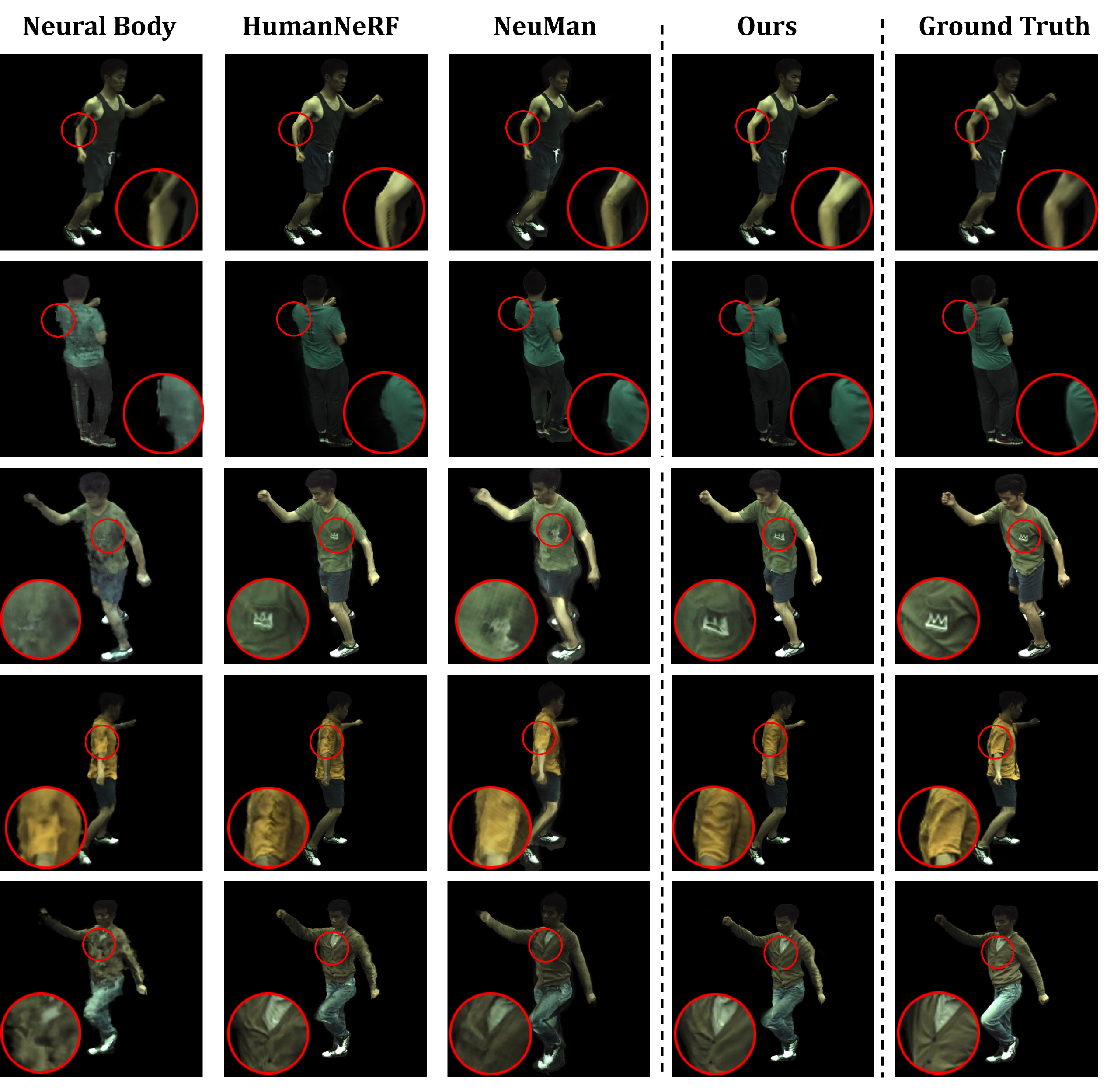}
    \vspace{-4mm}
    \caption{\textbf{Qualitative result of novel pose setting in ZJU-MoCap.} We compare the novel pose synthesis quality with baseline methods in ZJU-Mocap. Result shows that our method synthesizes more realistic images in novel poses.}
    \label{fig:np_zju}
\end{figure*}

To compare the generated results, we visualize the novel
pose synthesis results in ZJU-MoCap dataset of Neural-
Body, HumanNeRF and our method in Fig.~\ref{fig:np_zju}. NeuralBody
tends to generate vague images with large noise in novel
poses. HumanNeRF tends to produce some black line artifacts in some detailed areas. We also show the extract comparison with HumanNeRF under the challenge poses generated by MDM~\cite{tevet2022human} model in Fig.~\ref{fig:mdm_poses}. The result shows that our methods can retain these detail due to the correct deform and help with guided features.

We show the detailed quantitative results of each subject we compare in ZJU-MoCap dataset in Table.~\ref{table:detail_nv} and Table.~\ref{table:detail_np}. Though the PSNR metric of NeuralBody seems good in value, they synthesize poor visual quality images in both novel view and novel pose (as PSNR prefers smooth results). Our improvement in PSNR over HumanNeRF is not significant and slightly lower in subject 387 when testing in the novel pose.  But in LIPIS, our method has more significant improvement in novel view and novel pose setting respectively. We can see that our MonoHuman framework outperforms existing methods in most metrics in both settings.

\begin{figure*}[!t]
    \vspace{-2mm}
    \includegraphics[width=0.96\linewidth]{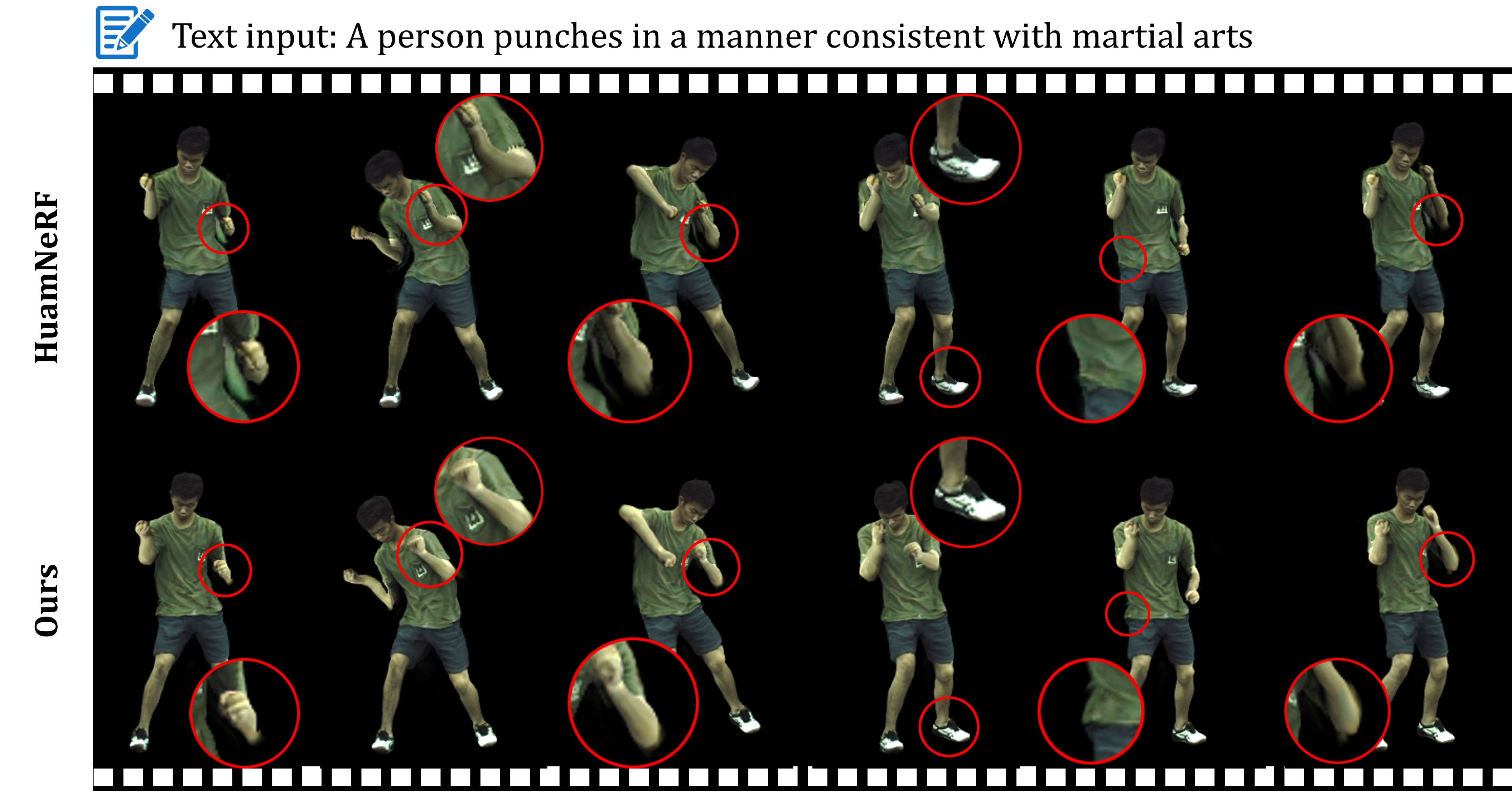}
    \vspace{-2mm}
    \caption{\textbf{Qualitative results on challenge poses generated by MDM~\cite{tevet2022human} through text input.} We evaluate our method driven by challenge pose sequence generated by MDM model.}
    \label{fig:mdm_poses}
\end{figure*}

\end{document}